\documentclass[conference]{IEEEtran}
\IEEEoverridecommandlockouts

\usepackage{cite}
\usepackage{tabularx}
\usepackage{url}
\usepackage[hidelinks]{hyperref}
\usepackage{booktabs}
\usepackage{amsmath,amssymb,amsfonts}
\usepackage{algorithmic}
\usepackage{graphicx}
\usepackage{textcomp}
\usepackage{soul}
\usepackage{xcolor}
\usepackage{array} 
\newcolumntype{Y}{>{\centering\arraybackslash}X}

\def\BibTeX{{\rm B\kern-.05em{\sc i\kern-.025em b}\kern-.08em
    T\kern-.1667em\lower.7ex\hbox{E}\kern-.125emX}}
\begin{document}

\title{Adversarially-Refined VQ-GAN with Dense Motion Tokenization for Spatio-Temporal Heatmaps
}


\author{
\IEEEauthorblockN{Gabriel Maldonado\IEEEauthorrefmark{1}, Narges Rashvand\IEEEauthorrefmark{1}, Armin Danesh Pazho\IEEEauthorrefmark{1}, Ghazal Alinezhad Noghre\IEEEauthorrefmark{1},\\ Vinit Katariya\IEEEauthorrefmark{2}, and Hamed Tabkhi\IEEEauthorrefmark{1}}
\IEEEauthorblockA{\IEEEauthorrefmark{1}University of North Carolina at Charlotte,
Charlotte, NC, USA\\
\{gmaldon2, nrashvan, adaneshp, galinezh, htabkhiv\}@charlotte.edu}
\IEEEauthorblockA{\IEEEauthorrefmark{2}University of Wyoming,
Laramie, WY, USA,
vkatariy@uwyo.edu}
\thanks{This research is supported by the National Science Foundation (NSF) under Award Numbers 2329816 and 1831795.}

}

\maketitle

\begin{abstract}
Continuous human motion understanding remains a core challenge in computer vision due to its high dimensionality and inherent redundancy. Efficient compression and representation are crucial for analyzing complex motion dynamics. In this work, we introduce an adversarially-refined VQ-GAN framework with dense motion tokenization for compressing spatio-temporal heatmaps while preserving the fine-grained traces of human motion. Our approach combines dense motion tokenization with adversarial refinement, which eliminates reconstruction artifacts like motion smearing and temporal misalignment observed in non-adversarial baselines. Our experiments on the CMU Panoptic dataset \cite{b21} provide conclusive evidence of our method's superiority, outperforming the dVAE baseline by 9.31\% SSIM and reducing temporal instability by 37.1\%. Furthermore, our dense tokenization strategy enables a novel analysis of motion complexity, revealing that 2D motion
can be optimally represented with a compact 128-token vocabulary, while 3D motion's complexity demands a much larger 1024-token codebook for faithful reconstruction. These results establish practical deployment feasibility across diverse motion analysis applications. The code base for this work is available at
\href{https://github.com/TeCSAR-UNCC/Pose-Quantization}{https://github.com/TeCSAR-UNCC/Pose-Quantization}.
\end{abstract}


\section{Introduction}


Human motion understanding is a critical area of research with applications spanning healthcare, robotics, and human-computer interaction. Advances in deep learning across diverse domains highlight the effectiveness of representation learning in extracting meaningful patterns from complex data \cite{b39, b40,b41,b42, b143}. A key challenge in human motion analysis is the effective representation of motion, which is inherently high-dimensional and complex \cite{b55, b56, b57, b54}. Among various representations, dense spatio-temporal heatmaps have emerged as particularly powerful, retaining rich spatial relationships that are often lost in sparse keypoint or skeleton-based formats \cite{b38, b239}. The effectiveness of volumetric heatmaps for spatiotemporal feature learning has been well-established \cite{b12}, but their high data footprint, coupled with potential noise and redundancy from recording equipment \cite{b49, b21}, pose a significant challenge for efficient analysis.

\begin{figure}[t]
    \centering
    \includegraphics[width=0.49\textwidth]{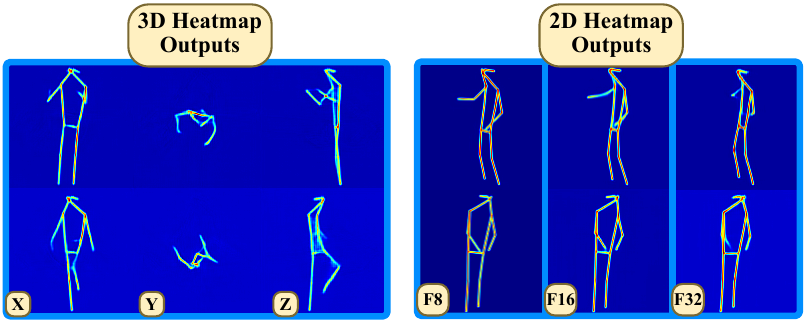}
    \caption{Reconstructed human poses from our VQ-GAN models, with 3D heatmaps across the X, Y, and Z on the left and 2D heatmaps at compression levels F8, F16, and F32 on the right.}
    \label{fig:examples}
\end{figure}

State-of-the-art generative models, including Variational Autoencoders (VAEs) \cite{b35, b43, chang2022maskgit} and Vector-Quantized GANs (VQ-GANs) \cite{b17, b18}, offer a promising paradigm for compressing this data into discrete tokens. However, their direct application to motion sequences reveals a critical flaw: standard reconstruction-based objectives are insufficient for preserving motion fidelity. Non-adversarial frameworks often produce visually convincing, but temporally inconsistent results, such as motion smearing and frame misalignment. This temporal degradation corrupts the fundamental motion patterns that downstream tasks like high-precision pose estimation \cite{b2,b3}, action recognition \cite{b47,b60}, and anomaly detection \cite{b50, b60, b57} rely on. While significant research has focused on text-to-motion synthesis using VQ-based models \cite{b37, b388}, the fundamental problem of high-fidelity, high-compression representation for motion analysis remains largely unaddressed.

To overcome these limitations, we introduce a novel Adversarially-Refined VQ-GAN framework, specifically engineered for the compression of dense spatio-temporal heatmaps. Our core contribution is the integration of an adversarial training objective that explicitly penalizes temporally unrealistic reconstructions, a component we demonstrate is essential for eliminating motion artifacts. While our architecture establishes a new state-of-the-art, we acknowledge that more expressive encoders, such as transformer-based designs \cite{b7, b8,b15}, represent a promising avenue for future enhancement. As illustrated in Figure~\ref{fig:examples}, our current framework maintains high-quality reconstructions even at aggressive compression factors, establishing a foundational method for motion analysis. To validate our framework, we conduct a series of comprehensive experiments on the large-scale CMU Panoptic dataset. Our analysis systematically investigates the trade-offs between compression fidelity and codebook size, directly comparing our adversarially-refined model against strong non-adversarial baselines like the discrete Variational Autoencoder (dVAE). Furthermore, we provide a detailed analysis distinguishing the representational requirements for 2D and 3D motion, revealing the impact of dimensionality on the learned discrete vocabulary. Through both quantitative metrics, such as the Structural Similarity Index (SSIM) and Temporal Standard Deviation (T-Std), and qualitative visualizations, we demonstrate the critical role of the adversarial objective in eliminating temporal artifacts and preserving the fine-grained traces of human motion across various compression levels. Our primary contributions can be summarized as follows:

\begin{itemize} 
    \item \textbf{An Adversarially-Refined VQ-GAN Framework for Motion Representation}: We introduce the first VQ-GAN framework for encoding human motion, discretizing both 2D and 3D spatio-temporal heatmap sequences into compact latent tokens. This generative architecture uses an adversarial objective to ensure temporal coherence, eliminating motion smearing and misalignment artifacts common in non-adversarial baselines.

    
    \item \textbf{Dense Motion Tokenization and Analysis}: We propose a novel dense tokenization strategy that captures subtle motion patterns often lost in sparse representations. By systematically analyzing the role of compression factors and vocabulary sizes, we provide a unique insight into the intrinsic complexity of motion. We demonstrate that 2D motion can be optimally represented with a compact 128-token vocabulary, whereas 3D motion demands a much larger codebook.

    \item
\textbf{Superior Compression and Fidelity Performance}: Our framework establishes a new state-of-the-art for motion compression. We demonstrate that our adversarially-enhanced discrete embeddings outperform dVAE models in both reconstruction quality and temporal stability. This confirms the viability of using discrete tokenization for demanding motion analysis tasks.

\end{itemize}

\section{Related Works}

Understanding human-centric environments requires a precise estimate of human body movements and poses. Human motion analysis has evolved significantly with deep learning, moving from handcrafted features to leveraging sophisticated Convolutional Neural Networks (CNNs) and Transformer-based architectures \cite{b2, b3, b24, b27}. These methods leverage Graph Convolutional Networks (GCNs) and spatio-temporal transformers \cite{b7, b5} to effectively model relationships between body joints over time, achieving state-of-the-art performance in 3D human pose estimation\cite{b49}.
More recently, research has focused on learning discrete latent representations for generative modeling. Discrete Variational Autoencoders (dVAEs) \cite{kingma2014} have shown promise for tasks like image and text-to-motion generation\cite{b19}, but they face challenges with non-differentiability and hyperparameter sensitivity \cite{b4, b32}. Vector-Quantized Variational Autoencoders (VQ-VAEs) \cite{b31} offer an alternative, using vector quantization to avoid these issues. VQ-VAEs have been successfully extended to human motion generation as seen in works like \cite{b35, b43, b48}. 
Building on the success of VQ-VAE, VQ-GAN \cite{b18} were developed for high-resolution image synthesis, combining the strengths of VAEs and GANs. While VQ-GANs excel in image generation \cite{b18, yu2022vqgan}, their application to human motion remains largely unexplored. Specifically, no prior work has leveraged VQ-GANs to encode and reconstruct spatio-temporal heatmap sequences, leaving a gap in compact yet interpretable motion representation learning. 

 To bridge this gap, we present our adversarially-refined VQ-GAN for human motion. 
 To contextualize our contribution, we provide a comparison of our work with other relevant methods in Table \ref{prior_work}. While works like MoMask\cite{b38} and MotionGPT \cite{b27} focus on generative tasks such as text-to-motion, our approach is distinct as it prioritizes high-fidelity motion compression and representation, making it more analogous to dVAE baselines. This is because both our method and dVAE focus on encoding and discretizing motion representations from heatmaps. However, our work distinguishes itself by uniquely applying a VQ-GAN-based approach to this problem, offering significant advantages in compression and a novel denoising (DN) capability not present in the other methods.

\begin{table*}[htp]

\centering
\caption{Comparison of recent motion synthesis methods with respect to input modalities, architectural designs, denoising capabilities, and application domains.}
\label{prior_work}
\begin{tabularx}{0.9\textwidth}{@{}c||Y|Y|Y|Y@{}}
    \toprule
    \textbf{Method} & \textbf{Input} & \textbf{Core Architecture} & \textbf{Denoising} & \textbf{Application} \\ 
    \midrule \midrule
    dVAE [50] & Spatio-temporal heatmaps & Transformer + dVAE & No & A foundation model for a wide range of motion-based tasks. \\ 
    \midrule
    MoMask\cite{b38} & Textual descriptions for motion generation & Masked transformer + residual vector quantization (RVQ) & No & Text-driven 3D human motion generation. \\ 
    \midrule
    MotionGPT\cite{b27} & Motion and text & Pre-trained LLM + VQ-VAE & No & Text-to-motion generation, motion captioning. \\ 
    \midrule
    Ours & 2D and 3D Heatmaps & VQ-GAN & Yes & Motion representation optimization and analysis. \\ 
    
    \bottomrule

\end{tabularx}
\end{table*}

\section{Methodology}
\begin{figure*}[h]
    \centering
    \includegraphics[width=0.85\textwidth]{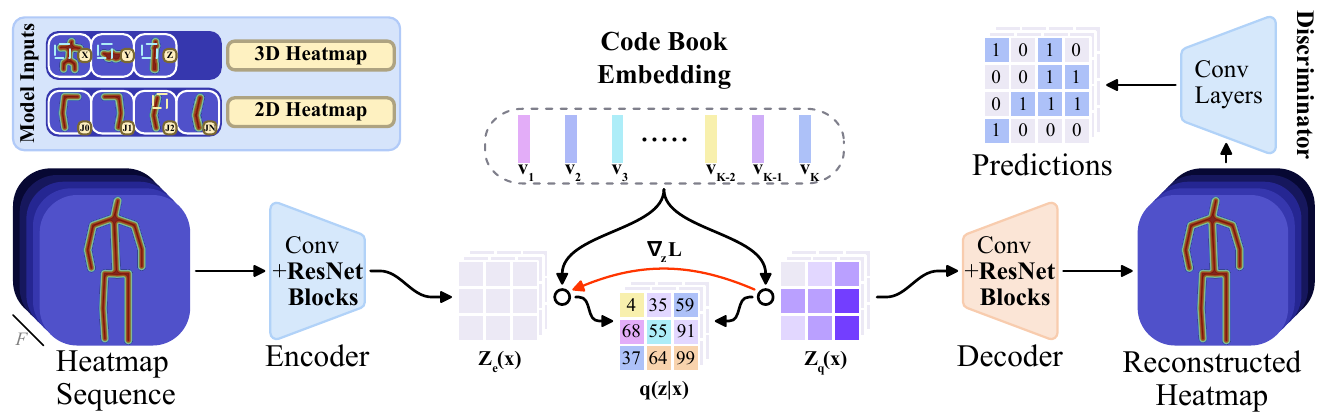}
    \caption{Our VQ-GAN model for motion representation takes 2D and 3D heatmap sequences as input, encodes them into a continuous latent space via convolutional and ResNet blocks, discretizes these features via codebook quantization, and reconstructs the heatmaps with a decoder, while a discriminator encourages realistic outputs by minimizing adversarial loss.}
    \label{fig:vq-gan_overview}
\end{figure*}

Our methodology is centered on learning an efficient and interpretable representation of human motion by employing a Vector Quantized Generative Adversarial Network (VQ-GAN). The proposed architecture, illustrated in Figure~\ref{fig:vq-gan_overview}, processes spatio-temporal heatmap sequences, which serve as a dense, image-like representation of both 2D and 3D pose keypoint data. 
In the following subsections, we detail each component, beginning with motion heatmap generation, followed by the VQ-GAN architecture and its adversarial refinement.

\subsection{Motion Heatmap Generation}
Our approach begins by transforming the raw, unstructured pose keypoint data into a structured format suitable for our deep learning model. We convert the coordinate-based keypoints into a sequence of spatio-temporal heatmaps that represent both 2D and 3D human pose. This crucial conversion provides a dense, image-like representation of motion, enabling our subsequent convolutional layers to efficiently extract localized features and capture the inherent spatial and temporal relationships within the data. The structure of these generated heatmaps is designed to fit the dimensionality of the input data as follows:

\begin{itemize}
    \item \textbf{2D keypoint inputs}: Raw 2D keypoints are sequence of $(x,y)$ coordinates pairs with  dimensionality \( P\in \mathbb{R}^{F\times K\times2} \), where $F$ is the number of frames, $K$ is the number of keypoints per frame. To provide an image-like representation that better encodes spatial context, keypoints are transformed into Gaussian-distributed heatmaps \(V\in\mathbb{R}^{F \times K\times H \times W} \), where \( F \) is the number of frames, \( K \) is the number of keypoint pairs, and \( H \times W \) represents the spatial dimensions. This captures temporal evolution of motion by mapping keypoint connections across frames, a method similar to \cite{b12}.

    \item \textbf{3D keypoint inputs}: 3D inputs are $(x,y,z)$ coordinates with dimensionality of \( P\in \mathbb{R}^{F\times K\times3} \).
    Heatmaps are represented as   \( V \in \mathbb{R}^{F \times K\times D \times H \times W} \), where \( D \times H \times W \) specifies the spatial resolution of each heatmap. These volumes are subsequently projected along the X-Y, Y-Z, and X-Z planes, which reduces dimensionality while retaining key spatial features for processing by the model's architecture.
\end{itemize}

The Gaussian functions for heatmap generation are defined as $U_{f2}$, which represent the value at a specific pixel. Here, $(x_j,y_j)$ is the keypoint coordinate, $(i,j)$ is the heatmap pixel coordinate, and $\sigma$ is a standard deviation that controls the spread of the Gaussian.
\begin{equation}
U_{f2} = e^{-\frac{(i - x_j)^2 + (j - y_j)^2}{2\sigma^2}}.
\end{equation}
For 3D keypoints, this Gaussian is extended to include the \(z\) dimension.


\subsection{VQ-GAN Architecture} 
\label{section:vqgan_architecture}
\begin{figure*}[ht]
    \centering
    \includegraphics[width=0.88\textwidth]{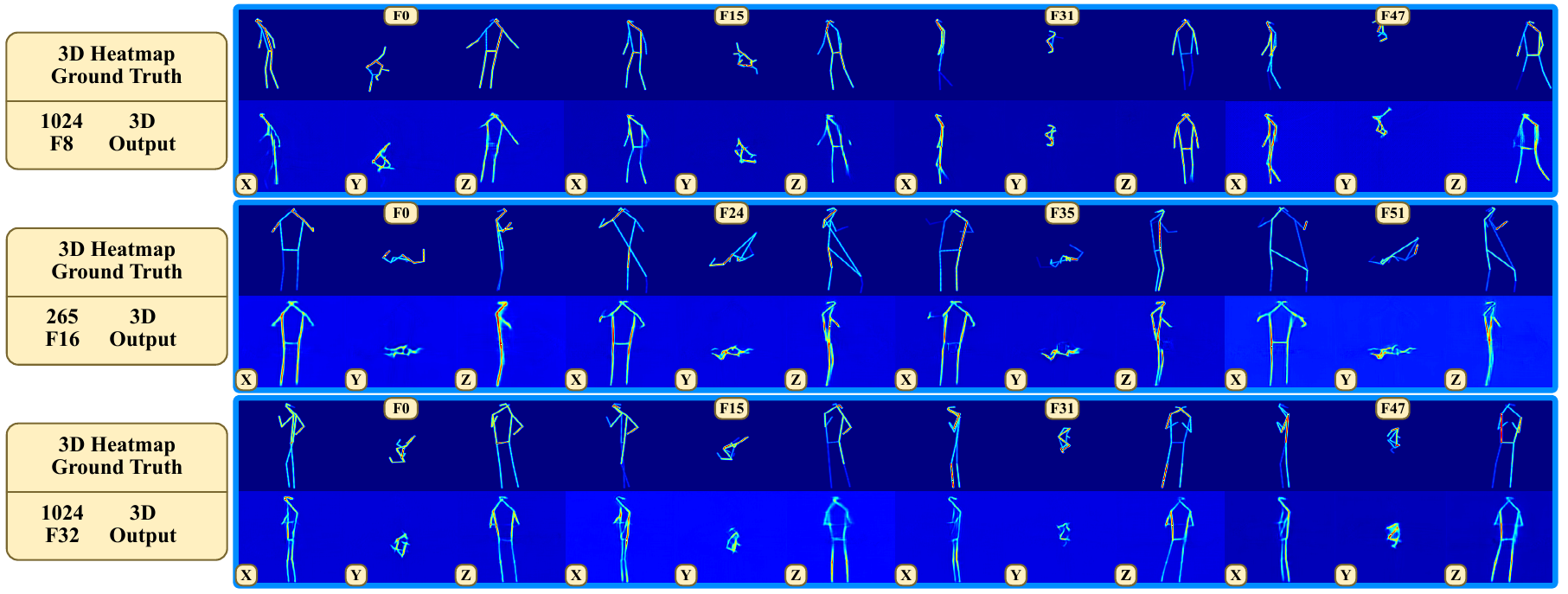}
    \caption{3D heatmap reconstructions under varying compression rates (F8, F16, F32)  and codebook sizes (1024 and 256), comparing ground-truth heatmaps with model outputs across motion sequences to demonstrate accurate motion representation even at high compression.}
    \label{fig:3D_proj_examples}
\end{figure*}

Once the spatio-temporal heatmap sequences are generated, they are processed through our Vector Quantized Generative Adversarial Network (VQ-GAN), illustrated in Figure~\ref{fig:vq-gan_overview}. The VQ-GAN architecture is designed with two primary objectives: first, to encode spatial and temporal information into a compact latent space, aiming to reduce the amount of information accepted while retaining structural and motion-related details; second, to discretize this information for use by the decoder in reconstructing the input provided to the encoder. The process compresses high-dimensional data into a structured, lower-dimensional latent space.
Our VQ-GAN architecture  combines autoencoding, vector quantization, and adversarial learning, as detailed below.

\subsubsection{Encoder}
The encoder is responsible for extracting meaningful spatio-temporal features from the motion heatmaps. It utilizes several layers of a 3D convolutional neural network, followed by adapted ResNet 3D convolutional blocks \cite{b22}, to extract key features from heatmaps. This is followed by a downsampling block to take the previously extracted information and reduce the size of the data. This hierarchical design ensures that the encoder produces a compact yet expressive latent representation that is well-suited for subsequent vector quantization.

\subsubsection{Vector Quantization and Codebook}
Instead of directly passing the continuous latent vectors forward, the encoder outputs are mapped to a learned codebook of embeddings. This vector quantization step discretizes continuous features into a finite set of tokens. By forcing the model to use a limited vocabulary of latent codes, the codebook acts as a form of regularization that promotes efficient encoding of motion patterns. Formally, Vector quantization (VQ) transforms the continuous latent representation produced by the encoder into a structured, discrete form. The encoder outputs continuous latent vectors $z_e$, which are mapped through \(q(z|x)\) to the nearest entry in a learnable codebook, yielding discrete vectors $z_q$. Each embedding captures relevant motion information, reducing redundancy and representing motion characteristics as indices within the Code Book. The Code Book contains a set of discrete vectors, which corresponds to the classifications generated by the model. This discretization increases the attention of the embeddings in vision transformers as shown in \cite{b19}.

\subsubsection{Decoder and Adversarial Training}
The decoder is responsible for reconstructing motion heatmaps from the quantized representations produced by the encoder and codebook. In the context of our VQ-GAN architecture, the decoder serves as the generator (G) of the GAN component. During training, the decoder maps the discrete codebook embeddings $z_q$ back into the heatmap space. This process is guided by a reconstruction loss, which enforces similarity between the reconstructed heatmaps and the ground-truth heatmaps. In addition, a discriminator network introduces an adversarial loss, encouraging the reconstructed heatmaps to be more realistic and closely match the ground truth.

\subsubsection{Optimization Objective}
 The total loss for our VQ-GAN model is a combination of three key terms, including reconstruction, quantization, and adversarial loss, which are jointly optimized to ensure the learned representation balances fidelity, compactness, and realism. This objective trains the encoder, vector quantizer, and decoder simultaneously. The overall loss can be expressed as: 
 
\begin{equation} \mathcal{L}_{Total} = \mathcal{L}_{\text{rec}} + \mathcal{L}_{\text{VQ}} + \lambda \mathcal{L}_{\text{adv}} \end{equation}
Here, $\mathcal{L}_{\text{rec}}$ is the reconstruction loss, $\mathcal{L}_{\text{VQ}}$ is the vector quantization loss, and $\mathcal{L}_{\text{adv}}$ is the adversarial loss. The hyperparameter $\lambda$ acts as a weighting factor to balance the adversarial contribution.

The reconstruction loss measures how well the decoder can recreate the original heatmap sequence from its quantized latent representation. To enforce both pixel-level accuracy and high-level perceptual quality, we define this loss as a weighted sum of a perceptual loss and a pixel-wise L1 loss.
\begin{equation}
L_{\text{rec}} = \alpha \cdot L_{\text{perceptual}} + \beta \cdot L_{\text{L1}} 
\end{equation}
The perceptual loss, $L_{\text{perceptual}} = ||\psi(x) - \psi(\hat{x})||_2$,
compares high-level feature representations of the ground-truth heatmap sequence $x$ and the reconstructed heatmap sequence $\hat x$ using a pre-trained VGG network. The $L_1$ loss, $L_{\text{L1}}$, provides a direct pixel-wise measure of reconstruction error.

$\mathcal{L}_{\text{VQ}}$ is the vector quantization loss that governs the process of mapping continuous latent vectors to discrete codebook entries.
\begin{equation}
L_{\text{VQ}} = ||\text{sg}[z_e(x)] - e||_2^2 + ||z_e(x) - \text{sg}[e]||_2^2
\end{equation}, where $z_e(x)$ is the continuous latent vector, $e$ is the selected discrete codebook vector that is closest to $z_e(x)$. The operator $\text{sg}[\cdot]$ denotes a stop-gradient, which prevents gradients from flowing back to a specified tensor during backpropagation.
The adversarial loss, $\mathcal{L}_{\text{adv}}$, is a key component for improving the visual quality and realism of the generated heatmaps. It is a min-max game between the VQ-GAN's decoder (the generator) and a discriminator. This loss has two parts: the generator's loss, $\mathcal{L}{G} = -\mathbb{E}{\hat{x} \sim p_G}[D(\hat{x})]$, and the discriminator loss, for which we use the hinge loss formulation. 
\begin{align}
\begin{split}
\mathcal{L}{D} &= \mathbb{E}{x \sim p_x}[\max(0, 1 - D(x))] \\
& + \mathbb{E}_{\hat{x} \sim p_G}[\max(0, 1 + D(\hat{x}))]
\end{split}
\end{align}, where $D(x)$ is the discriminator's output for a real heatmap and $D(\hat(x))$ is the output for a fake heatmap. This combined objective ensures the encoder and quantizer do not simply compress data arbitrarily, but instead capture motion features that remain faithful to the true dynamics of human movement.


\begin{figure}[ht]
    \centering
    \includegraphics[width=0.49\textwidth]{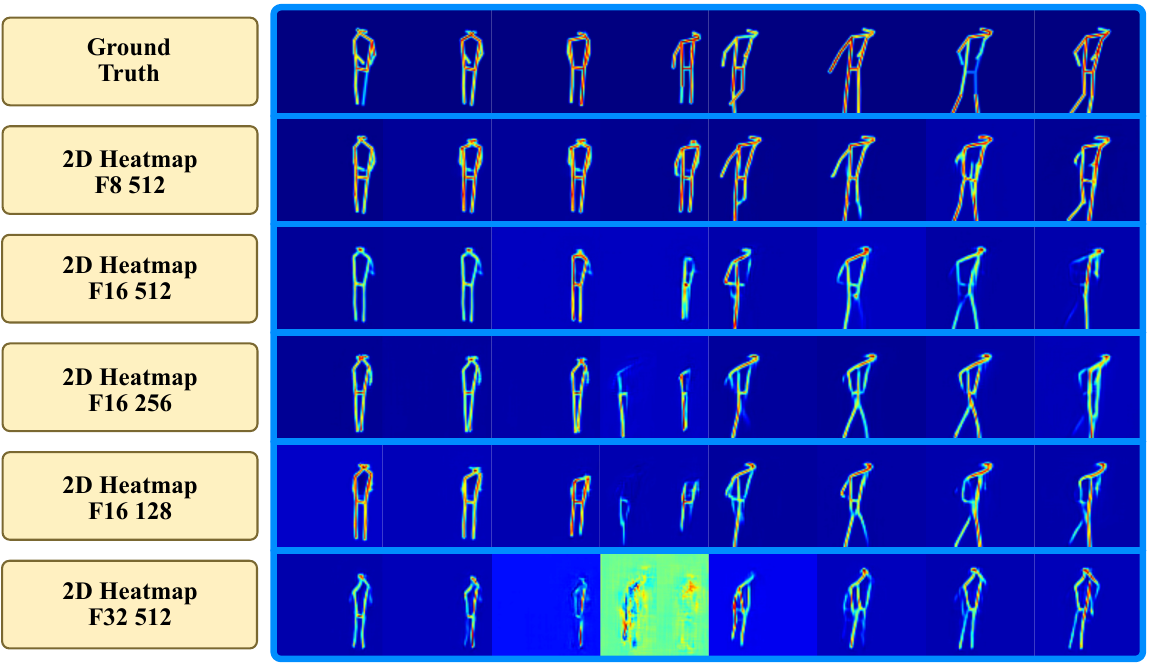}
    \caption{2D heatmap reconstructions for a single motion sequence under different compression rates (F8, F16, F32) and codebook sizes (512, 256, 128), illustrating the trade-off between efficiency and reconstruction quality.
    }
    \label{fig:multi_ego_examples}
\end{figure}

\begin{figure*}[h]
    \centering
    \includegraphics[width=0.866\textwidth]{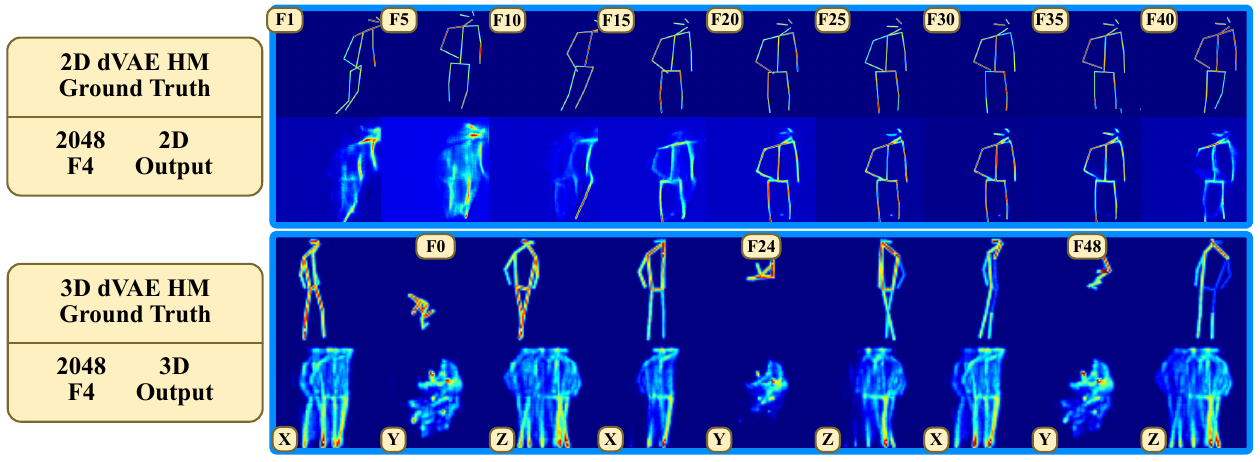}
    \caption{This figure shows reconstruction results of 2D and 3D dVAE models on human motion data. The top compares 2D ground-truth (top row) with reconstructed output (bottom row); the bottom does the same for 3D poses at frames 0, 24, and 48. Results demonstrate the model's ability to reconstruct both 2D and 3D poses under a compression factor of $F4$ and a codebook size of 2048.}
    \label{fig:dvae_output}
\end{figure*}

\section{Experimental Design}
\subsection{Dataset Selection}

Our work required a comprehensive dataset containing multi-view sequences of human pose information, spanning diverse actions and full ranges of motion to ensure generalization. To meet this requirement, we selected the CMU Panoptic dataset \cite{b21}. This dataset includes 31 high-definition (HD) cameras, supplemented by 480 VGA cameras with Kinect sensors, all positioned within a large dome facing inward to capture participants performing a wide variety of activities. From this setup, we selected 10 diverse cameras to provide different viewpoints. In total, we used 60 sessions (47 for training and 13 for validation). The training split contained 1,267,376 valid frames, which were segmented into 64-frame windows with a temporal stride of 90, yielding 14,081 motion segments. The validation split, drawn from a fixed set of sessions, including haggling interactions, musical performances, small group discussions, and office activities, produced 2,672 segments using the same procedure.



\subsection{Compression and Vocabulary Optimization} 
We analyze the effect of varying compression levels on the model’s ability to preserve essential motion information. To ensure fairness across settings, the number of codebook vectors is kept constant at a high value, allowing us to isolate the impact of compression alone. The compression factor is obtained by jointly reducing the temporal dimension ($T$) and the two spatial dimensions ($H$ and $W$) of the heatmap sequence.
\begin{table}[h!]
\centering
\caption{Compression settings showing original and reduced spatial–temporal dimensions, along with the resulting compression factors.}
\scalebox{0.89}{
    \begin{tabular}{@{}c|c|c|c@{}}
        \toprule
        \textbf{Compression Rate} & \textbf{Original Size} & \textbf{Compressed Size} & \textbf{Compression Factor} \\ \midrule \midrule
        {F8}               & 64$\times$128$\times$128             & 8$\times$16$\times$16                  & 512$\times$                         \\ 
       {F16}              & 64$\times$128$\times$128             & 4$\times$8$\times$8                    & 4096$\times$                        \\ 
        {F32}              & 64$\times$128$\times$128             & 2$\times$4$\times$4                    & 32,768$\times$                      \\ \bottomrule
    \end{tabular}}
    \label{tab:comp}
\end{table}

Table \ref{tab:comp} summarizes the relationship between the original input dimensions, the compressed size, and the corresponding compression factor for each setting.  As shown, the compression rate increases exponentially, resulting in a compression of up to 32,768 times of the original input size. This exponential reduction greatly decreases storage and computational demands while enabling a focused evaluation of how aggressively the model can compress the data without losing the spatio-temporal patterns essential for accurate heatmap reconstruction and downstream motion analysis.

We also assess different Code Book sizes to analyze the optimal hyperparameters for discretizing continuous human body motion. The objective is to identify an optimal Code Book size that maximizes representation capacity without overloading the model. This ensures that the discrete tokens generated by the VQ-GAN effectively encode motion dynamics. Throughout these evaluations, the compression rate remains consistent across all tested sizes.




\subsection{Metrics}
We evaluated our model using five complementary metrics: Structural Similarity Index (SSIM) \cite{b33}, Peak Signal-to-Noise Ratio (PSNR), L1 loss, Temporal Standard Deviation (T-Std), and quantization loss (Q-loss). Together, these capture both spatial fidelity and temporal stability, consistent with prior work in image quantization \cite{b34}. SSIM measures structural similarity between predicted and ground-truth heatmaps, \[
SSIM(x,y) = \frac{(2\mu_x\mu_y + C_1)(2\sigma_{xy}+C_2)}
{(\mu_x^2 + \mu_y^2 + C_1)(\sigma_x^2 + \sigma_y^2 + C_2)}
\], where $x$ and $y$ are the ground-truth and reconstructed heatmaps, $\mu_x$ and $\mu_y$ is the mean pixel values of $x$ and $y$, $\sigma$ is the variances of $x$ and $y$, $\sigma_{xy}$ is the covariance between $x$ and $y$, and $C_1$ and $C_2$ is the stabilizing constants. Additionally, PSNR reflects reconstruction distortion, 
\[PSNR = 10 \cdot \log_{10}\left(\frac{MAX_I^2}{MSE}\right) \], where $MAX_I$ is the maximum possible pixel value of the heatmap and $MSE$ is the Mean Squared Error between the reconstructed heatmap and the ground-truth. L1 loss provides the average pixel-wise error, \[L_1(x, y) = \frac{1}{N} \sum_{i=1}^{N} |x_i - y_i|\], where $x$, $y$ is the reconstructed and ground-truth heatmap, $N$ is the total number of pixels. T-Std  quantifies temporal smoothness by assessing pixel intensity variation across frames, {\small \[
    T\text{-}Std = \frac{1}{F \cdot H \cdot W} 
    \sum_{f=1}^{F} \sqrt{\frac{1}{H \cdot W} 
    \sum_{i=1}^{H}\sum_{j=1}^{W} 
    \left( I_{f}(i,j) - \mu_{f}(i,j) \right)^2 }
\] }, where $F$ is the number of frames, $H$ and $W$ are the height and width, $I_f(i,j)$ is the pixel intensity at position $(i,j)$ in frame $f$ and $\mu_f$ is the mean pixel intensity at position $(i,j)$, while Q-loss measures the quantization error between latent features and their codebook representations.

\subsection{Implementation Details}
We trained each model for 210,000 steps (15 epochs) on a single Nvidia A6000 GPU, which took approximately 2.25 days per session. We used the AdamW optimizer with a learning rate of $2.25 \times 10^{-5}$, $\beta_1 = 0.5$, and $\beta_2 = 0.9$. For evaluation, we adopted the discrete dVAE architecture from \cite{b31} to analyze the trade-off between compression and reconstruction quality. Our assessment combined qualitative visual inspection of motion fidelity with quantitative metrics for spatial accuracy, temporal stability, and quantization efficiency.

\section{Results}

To validate our framework, we conduct a comprehensive evaluation on the CMU Panoptic dataset \cite{b21}. Our analysis is presented as a qualitative assessment of the reconstructed motion to visually identify key differences between models, followed by a quantitative analysis that provides experimental evidence for our claims regarding temporal coherence, motion complexity, and compression fidelity.

\subsection{Qualitative Results}

Visual inspection of reconstructed motion sequences reveals differences in how adversarially-refined VQ-GAN and the dVAE baseline handle temporal dynamics.

Our VQ-GAN models, for both 2D and 3D, in Figures ~\ref{fig:multi_ego_examples} and ~\ref{fig:3D_proj_examples}, produce temporally coherent reconstructions even during rapid movement. They successfully preserve key motion characteristics across various compression rates, with minor, graceful degradation at higher levels like F16. While aggressive F32 compression introduces some noticeable structural loss, the core motion remains intelligible. The dVAE baseline in Figure~\ref{fig:dvae_output} struggles with temporal coherence. The reconstructions suffer from motion smearing, an artifact where the structure of limbs blurs across frames during movement, creating abrupt transitions. This visual evidence suggests a failure of the non-adversarial model to enforce realistic temporal constraints that we will quantify in the following section. This visual clarity confirms that the adversarial objective successfully removes the temporal artifacts inherent in non-adversarial approaches, preserving the fine-grained traces of human motion. Overall, the qualitative results demonstrate a difference in how the VQ-GAN and dVAE handle motion compression. The VQ-GAN's quantized representation effectively preserves both spatial and temporal coherence, while the dVAE's continuous latent space results in distinct artifacts based on each model's dimensionality.

\subsection{Quantitative Results}

\begin{table*}[htp]
\centering
\caption{Quantitative comparison of VQ-GAN and dVAE models, highlighting how VQ-GANs more effectively retain spatio-temporal structure and the trade-off between compression and quality.}
\scalebox{1.1}{
    \begin{tabular}{@{}l|c|c|ccccc@{}}
        \toprule
        \textbf{Model Type}        & \textbf{Compression}  & \textbf{Vocab Size} & \textbf{SSIM$\uparrow$} & \textbf{PSNR$\uparrow$} & \textbf{L1$\downarrow$}  & \textbf{T-Std$\downarrow$} & \textbf{Q-Loss$\downarrow$} \\ \midrule
        \textbf{2D VQ-GAN} & {F8}           & 512                 & 0.975      & 31.23     & 0.005     & 0.212     & 0.0013 \\ 
        \textbf{2D VQ-GAN} & {F16}          & 512                 & 0.950      & 28.06     & 0.008     & 0.217     & 0.0033 \\ 
        \textbf{2D VQ-GAN} & {F16}          & 256                 & 0.953      & 28.30     & 0.008     & 0.217     & 0.0034 \\
        \textbf{2D VQ-GAN} & {F16}          & 128                 & 0.954      & 28.39     & 0.007     & 0.219     & 0.0003 \\
        \textbf{2D VQ-GAN} & {F32}          & 512                 & 0.913      & 25.28     & 0.011     & 0.222     & 0.0003 \\ 
        \midrule 
      
        \textbf{2D dVAE (Baseline)}\cite{b130}     & {F8}          & 2048                & 0.925      & 27.07     & 0.010     & 0.2421    & --- \\
    
       \midrule 
        \textbf{3D VQ-GAN}     & {F8}           & 1024                & 0.934      & 31.65     & 0.005     & 0.151     & 0.0014 \\ 
        \textbf{3D VQ-GAN}     & {F16}          & 1024                & 0.912      & 28.45     & 0.008     & 0.237     & 0.0010 \\
        \textbf{3D VQ-GAN}     & {F16}          & 512                 & 0.891      & 26.95     & 0.009     & 0.235     & 0.0010 \\
        \textbf{3D VQ-GAN}     & {F16}          & 256                 & 0.866      & 27.21     & 0.009     & 0.219     & 0.0010 \\
        \textbf{3D VQ-GAN}     & {F32}          & 1024                & 0.858      & 26.53     & 0.011     & 0.225     & 0.0001 \\
        \midrule 
        \textbf{3D dVAE (Baseline)}\cite{b130}     & {F8}          & 2048                & 0.847      & 24.56     & 0.018     & 0.2399     & --- \\
        \bottomrule
    \end{tabular}
    }
    \label{tab:results}

\end{table*} 


Table \ref{tab:results} reports the quantitative performance of the proposed VQ-GAN models and the dVAE baselines across multiple compression levels and vocabulary sizes. Metrics capture both spatial fidelity (SSIM, PSNR, L1) and temporal stability (T-Std), as well as quantization efficiency (Q-Loss for VQ-GAN models). First, it can be observed that adversarial refinement is essential for temporal coherence. Comparing our best 3D VQ-GAN (F8) to the 3D dVAE baseline, our model achieves a 9.31\% higher SSIM (0.934 vs. 0.847). More critically, it reduces temporal instability, the quantitative measure of motion smearing, by 37.1\% (T-Std: 0.151 vs. 0.240). This empirical result provides strong backing for our qualitative observations, proving that the adversarial objective is directly responsible for eliminating the temporal artifacts that challenge non-adversarial models. Secondly, the dense tokenization strategy reveals a fundamental difference in the inherent complexity of 2D and 3D motion. For 2D models at F16 compression, reducing the vocabulary size from 512 to 128 surprisingly improves SSIM (0.950 to 0.954) while causing a 90.9\% collapse in quantization loss (0.0033 to 0.0003). This indicates that a compact 128-token codebook is not only sufficient but optimal for 2D motion, acting as a powerful regularizer. In contrast, the 3D models require a much larger vocabulary; reducing the codebook from 1024 to 512 at F16 results in a significant drop in fidelity. This finding, that 3D motion requires a codebook at least 8 times larger than 2D motion for high-fidelity reconstruction, is a novel insight into the dimensional complexity of motion. Finally, the results establish the practical viability of our framework for high-fidelity compression. At a moderate F16 compression, the model delivers excellent performance, achieving 95.4\% SSIM for 2D motion and 91.2\% SSIM for 3D motion. Even at an aggressive F32 compression, the model maintains high quality (0.913 SSIM for 2D), demonstrating graceful degradation. 

Across all configurations, our adversarially-refined VQ-GAN models consistently outperform their dVAE counterparts, confirming their superior ability to preserve spatio-temporal structure for real-world applications. The success of our VQ-GAN framework in creating a compact and semantically rich representation of human motion has significant implications for a wide range of downstream tasks where efficiency and high-level motion understanding are crucial. While our primary focus has been on high-fidelity motion reconstruction and compression, the discrete tokens produced by our model can serve as a foundational backbone for future work. Specifically, these tokens can be directly utilized in applications such as action classification. By replacing raw, high-dimensional video data with a sequence of motion tokens, a classifier can operate on a much smaller, yet information-rich input, potentially leading to faster inference and improved generalization. Similarly, the model's ability to capture fine-grained motion dynamics makes it an ideal foundation for anomaly detection. By pre-training on a large dataset of normal human behavior, the learned motion vocabulary can be used to identify anomalous or unusual motions as sequences that do not conform to the established codebook.

\section{Conclusion}

In this paper, we introduced an adversarially-refined VQ-GAN framework for the high-fidelity compression of dense spatio-temporal heatmaps, addressing the critical challenge of maintaining temporal coherence in motion reconstruction. Our results show that adversarial refinement removes motion smearing and improves temporal stability by over 37\% compared to dVAE baselines. The proposed dense motion tokenization further revealed that 2D motion can be optimally represented with a compact 128-token vocabulary, while 3D motion requires a much larger 1024-token codebook, offering a principled guide for designing efficient, dimension-aware compression models. By achieving high fidelity (over 95\% SSIM for 2D and 91\% for 3D) at practical compression rates, our work confirms the viability of using discrete tokenization for demanding motion analysis tasks. The compact nature of these learned tokens opens promising avenues for their direct use in future downstream applications, including action recognition, motion prediction, and anomaly detection, without the need for full reconstruction.

\end{document}